\documentclass[sigconf]{acmart}
\usepackage{multirow}
\usepackage{scalefnt}
\usepackage{smartdiagram}
\usepackage{tikz}
\usetikzlibrary{shapes.geometric, arrows}
\makeatletter
\newif\if@restonecol
\makeatother

\usepackage[linesnumbered,ruled,vlined]{algorithm2e}%
\usepackage{algpseudocode}
\renewcommand\footnotetextcopyrightpermission[1]{} %
\AtBeginDocument{%
  \providecommand\BibTeX{{%
    \normalfont B\kern-0.5em{\scshape i\kern-0.25em b}\kern-0.8em\TeX}}}

\begin{document}

\title{RLAD: Time Series Anomaly Detection through Reinforcement Learning and Active Learning}
\author{Tong Wu}
\affiliation{%
  \institution{Rutgers University}
}
\email{tw445@scarletmail.rutgers.edu}

\author{Jorge Ortiz}
\affiliation{%
  \institution{Rutgers University}
}
\email{jorge.ortiz@rutgers.edu}

\begin{abstract}
We introduce a new semi-supervised, time series anomaly detection algorithm that uses deep reinforcement learning (DRL) and active learning to efficiently learn and adapt to anomalies in real-world time series data.  Our model -- called RLAD -- makes no assumption about the underlying mechanism that produces the observation sequence and continuously adapts the detection model based on experience with anomalous patterns.  In addition, it requires no manual tuning of parameters and outperforms all state-of-art methods we compare with, both unsupervised and semi-supervised, across several figures of merit.  More specifically, we outperform the best unsupervised approach by a factor of 1.58 on the F1 score, with only 1\% of labels and up to $\sim$4.4x on another real-world dataset with only 0.1\% of labels.  We compare RLAD with seven deep-learning based algorithms across two common anomaly detection datasets with up to $\sim$3M data points and between 0.28\% -- 2.65\% anomalies.  We outperform \emph{all of them} across several important performance metrics.
\end{abstract}

\maketitle
\pagestyle{plain} %

\section{Introduction}

Time series anomaly detection is important across many application domains.  Data center monitoring systems~\cite{srcnn}, sensor networks~\cite{1824773} and the internet of things~\cite{devicemein}, cyber-physical systems~\cite{2984465,sbs} and finance~\cite{finance_AD}, among many others. 
However, most anomaly detection algorithms are tightly coupled to application-specific data properties.  Models are designed for a specific application; setting thresholds, feature extraction processing and hyper-parameters based on strong prior knowledge or extensive experimentation. Some properties include multi-scale temporal dependencies in a multivariate time series setting, anomalies in the frequency domain, and other associated statistical moments.  Deep learning techniques have recently been described in the literature.  For example SR-CNN~\cite{srcnn} casts time-series data into a visual context and spatial correlations are used to identify anomalies.  Recently, there has been a new, deep-learning based semi-supervised technique, Deep-SAD~\cite{ruff2019deep} which uses an information theoretic criteria to find anomalies.

In addition, anomaly detection algorithms must overcome two major challenges: 1) anomalies are rare and hence, models are difficult to train and 2) many real-world applications have non-stationary properties; that is, the underlying mechanism producing a series of measurements, changes over time.  In order to deal with the lack of labeled data, there have been many proposal in the literature that take an unsupervised approach~\cite{luminol,vallis2014novel,xu2018unsupervised,srcnn,siffer2017anomaly}.  However, unsupervised algorithms tend to perform quite poorly in practice and are significantly outperformed by semi and fully supervised methods~\cite{laptev2015generic, liu2015opprentice, ruff2019deep, pang2019deep}.  Supervised models perform very well when labels are available, however do not perform well when there are few or no labels available.  Even when labels are readily available, they fundamentally assume that the underlying distribution is stationary.  If the distribution changes, they must be re-trained.

To address these challenges, we must design an algorithm that can learn efficiently, using only a fraction of the labels necessary to train a supervised model.  To address the second challenge, we must consider adaptive designs; algorithms that adapt to changes as they learn more about the data and the distribution of anomalies within it.  We introduce a new algorithm called RLAD, combines reinforcement learning with active learning and label propagation to learn an anomaly-detection model that generalizes and that can adapt and improve dynamically, as it observes more data and learns more about anomalous samples.  We show that our model quickly learns an effective anomaly detection model using very few labeled examples,  outperforming several state-of-the-art, unsupervised and semi-supervised techniques.
We run extensive experiments on several public datasets and show superior performance over the competing state-of-art methods.

To the best of our knowledge, we are the first to combine deep reinforcement learning for anomaly detection with active learning.  There are techniques that use reinforcement learning for anomaly detection~\cite{huang2018towards} but are difficult to train without a significant number of labeled samples.  There are also techniques that use active learning, showing competitive performance to semi-supervised models with a fraction of the labels~\cite{zhang2019cross}.  Our contributions are as follows:
\begin{itemize}
    \item We demonstrate the first combination of deep reinforcement learning and active learning for anomaly detection, RLAD.
    \item We run extensive experiments on 3 real world anomaly data sets and 7 semi-supervised and unsupervised anomaly detection algorithms.
    \item We show that RLAD outperforms unsupervised techniques by a large margin ($\sim$4.4x the F1 score on some data sets)
    \item We outperform the state-of-the-art, semi-supervised technique Deep-SAD~\cite{ruff2019deep} by up to $\sim$10x in our experiments.
\end{itemize}

The the rest of the paper we discussed some related work in section \ref{related_work} and introduce preliminary knowledge in section \ref{preliminaries}. In section \ref{system_overview}, we present an overview of our anomaly detection system. Next, we give a more detailed description of our approach in section \ref{methodology}. In section \ref{experiment}, we provide the experiments of RLAD and evaluation with other approaches. We conclude our work in section \ref{conclusion}.

\section{Related work}\label{related_work}
In this section, we review several related works but we focus on deep-learning based methods, since they are the most relevant to RLAD.

\subsection{Traditional approach}
    Most traditional approaches are usually simple machine learning algorithms based on distance and density such as K-nearest-neighborhood(KNN)~\cite{angiulli2002fast}, local outlier factor(LOF)~\cite{breunig2000lof}, local correlation integral(LOCI)~\cite{papadimitriou2003loci}, isolation forest(iForest)~\cite{liu2012isolation}. And others like one-class support vector machine(OCSVM)~\cite{platt1999estimating}, principle component analysis(PCA)~\cite{shyu2003novel} are commonly used in this area. These approaches are usually time efficient but not able to achieve high performance in real practice.
\subsection{Supervised approach}
Recently, deep learning has achieved much success in anomaly detection area. Most deep learning based approaches usually extract features from data and build a classification model to identify anomalies. Supervised deep anomaly detection involves training a deep supervised binary or multi-class classifier, using labels of both normal and anomalous data instances. For instance,  ~\cite{laptev2015generic} is the anomaly detection system of Yahoo, called EGADS. It uses a collection of anomaly
detection and forecasting models with an anomaly filtering
layer for accurate and scalable anomaly detection on timeseries.
There are also general anomaly detection supervised algorithms which are not specific for time series but can be applied to this problem. Opperentice \cite{liu2015opprentice} used operators’ periodical labels on anomalies to train a random forest classifier and automatically select parameters and thresholds. However, these methods rely on tons of labels to train their models, which is quite hard to get in real world. 

\subsection{Unsupervised approach}
Advanced unsupervised learning techniques become more popular in this area. They do not require labels and assume the abnormal points have larger deviation from the normal distribution. LinkedIn developed their anomaly detection system Luminol\cite{luminol}, which is a light library of Python. Twitter proposed TwitterAD \cite{vallis2014novel} which is written in R and employs statistical learning to detect anomalies in both application as well as system metrics. Xu, et al. proposed DONUT ~\cite{xu2018unsupervised}, an unsupervised anomaly detection system based on Variational Auto ENcoder(VAE).
SR-CNN from Microsoft and Luminol from LinkedIn are proposed to deal with this problem. SR-CNN \cite{srcnn} from Microsoft which borrows the SR model from visual saliency detection domain to time-series anomaly detection and combines with CNN to improve performance. In 2017, SOPT and DSPOT are proposed \cite{siffer2017anomaly} on the basis of Extreme Value Theory\cite{de2007extreme}. DAGMM \cite{zong2018deep} utilizes a deep autoencoder to generate a low-dimensional representation and
reconstruction error for each input data point, which is further fed into a Gaussian Mixture Model. However, the performance achieved by these is rather low\cite{gornitz2013toward} since they are difficult to leverage prior knowledge (e.g., a few labeled anomalies) when such information is available as in many real-world anomaly detection applications or there is availability of human expert. Moreover, they require the assumption that normal pattern is stationary and are not able to adapt to shifts in input distributions. 

\subsection{Semi-supervised approach}
Semi-supervised approaches use a fraction of labeled data for training. They are commonly used in classification \cite{kingma2014semi, rasmus2015semi, dai2017good,oliver2018realistic}. Only a few of them have been proposed for anomaly detection problem. \cite{mcglohon2009snare} and \cite{tamersoy2014guilt} used label propagation process to achieve anomaly detection but only applicable for graph data. \cite{pang2018learning} introduced REPEN which leverages a few labeled anomalies to learn more relevant features. Deep-SAD \cite{ruff2019deep} used  an information-theoretic perspective on anomaly detection and an autoencoder model for pre-training. It uses a hyper-parameter to control the balance of training data. \cite{pang2019deep} proposed a neural deviation learning based approach and leveraged a few labeled anomalies with a prior probability to enforce the deviation of anomalies in normal distribution. However, these methods highly rely on the number of labeled anomalies in training data. But anomalies are rare in anomaly detection problem. Given a set of unknown samples,  it has to inspect a large amount of unknown samples to filter required amount of anomalies. It makes labels more expensive.

\subsection{Reinforcement learning based approach}
In the past three years, reinforcement learning attempted to be introduced in anomaly detection. It could solve problem that sequences without a clear normal pattern because of its self-improving characteristic. \cite{huang2018towards} proposed an idea to apply deep reinforcement learning on time-series anomaly detection. However, the performance is not able to satisfy real world applications and fully-labeled data is required. \cite{oh2019sequential} applied inverse reinforcement learning technique but it is not specified for time series and need fully-labeled data.

\section{Preliminaries} \label{preliminaries}
\subsection{Reinforcement learning}
We consider our anomaly detection problem as a Markov decision process(MDP) which could be expressed by a tuple of $<S, A, P_a, R_a, \gamma>$. $S$ represents the set of environment states. $A$ is the set of actions taken by RL agent. $P_a(s, s')$ means the probability that action $a$ is performed at the state $s$ and will lead to state $s'$. $R_a(s, s')$ is the instant reward received by agent from taking the action $a$. $\gamma  \in [0,1]$ is the discount factor.  The value function is defined as $V_{\pi}(s) = \mathbb{E}[\sum_{t}^{T_{\infty}} \gamma^t R_{t}|s_0 = s]$ represents the expected reward return starting from state $s$. In MDP, the agent is trying to learn a control policy $\pi : S \rightarrow A$ that maximizes the accumulated future reward $\sum_{t}^{T_{\infty}} \gamma^t R_{t}$.

Typically, the proposed methods to solve Markov decision process problems are model-based methods and model-free methods. Model-based methods build a model to represent the environment. The agent has to learn and understand the environment to interact with it. The most common model-based algorithm is dynamic programming.  It is a method for solving complex problems by breaking them down into sub-problems. Model-free methods do not try to get a full understanding of environment but just simply explore environment, imagine the next situation and take the best actions he believes. Since model-free algorithms have more general applications and usages, we only talk about model-free algorithms in this paper. Two classes of model-free algorithms are proposed: value-based algorithm and policy-based algorithm.

\subsection{Deep Q Network}
Q learning is a famous value-based algorithm. Agents learns the action-value function $Q(s,a)$ and predicts how good to take an action as a specific state. The target value is defined by:
$$
\text {target}=R\left(s, a, s^{\prime}\right)+\gamma \max _{a^{\prime}} Q_{k}\left(s^{\prime}, a^{\prime}\right)
$$
And the $Q$ function is updated by:
$$
Q_{k+1}(s, a) \leftarrow(1-\alpha) Q_{k}(s, a)+\alpha\operatorname{target}\left(s^{\prime}\right).
$$

Unfortunately, Q learning is unstable or even divergent when action value function is
approximated with a nonlinear function like neural networks.\cite{li2017deep} Thus, a method called  Deep Q network(DQN) that combines reinforcement learning with deep neural network is proposed by \emph{Deepmind} and approved to be able to handle more complicated problems.  It stabilizes the training of action value function approximation into a deep neural network. The key elements of DQN are experience replay\cite{replay} and target network.

Experience replay stores the transitions of tuple $<s, a, r, s'>$ where $s$ is defined by state, $a$ represents the action, $r$ is the reward by taking action $a$ and $s'$ is the next state after taking action. The agent is trained by randomly sampling a mini batch from it during each iteration. It reduces the correlation between samples and improves data efficiency. 

Target network is designed to deal with non-stationary target value problem. We build a neural network to learn values of $Q$ but target values are really unstable, which makes training difficult to converge. Therefore, another neural network which has a exact same structure called target network is designed to fix target values temporarily in order to speed training process.

\subsection{Active learning}
Active learning is a specific machine learning algorithm which allows the model actively interacts with user to obtain the desired learning experience. The machine(learner) could ask the desired data used in learning process. 

Assume we have a training set $L = (X, Y)$ and an unlabeled instance pool $U = (x_1, x_2, \dots, x_n)$. The unlabeled data is normally cheaper than labeled data. Thus, the $U$ can be very large. Active learning is going to select the unlabeled samples from $U$ and ask a human expert to label them manually through query function $Q$. Typically, the chosen samples have more valuable information to the current model $C$. The learner will get most training benefits from the new labeled instances. The new labeled instance set $L_{new} = (X_{new}, Y_{new})$ will be added to the training set $P$ and used for training in the next iteration. The goal of active learning is to achieve a good classifier model $C$ with a reasonable number of queries.

There are multiple types of query strategies according to what kind of data the learner want acquire. Typically, the general and most commonly used strategies are:
\begin{itemize}
    \item Random select: Randomly select samples from $U$ and throw into $L$;
    \item Least confidence: Select the most uncertain samples $x_{lc}$ for the classifier $C$ to identify;
    \begin{equation}
        x_{lc} = \arg  \max{1-P_{C}(\hat{y}|x)}
    \end{equation}
    where $\hat{y} = \arg \max_{y}{P_{C}(y|x)}$ or the the class label with the highest posterior probability under the model $C$\cite{settles2009active}. This shows the model lack of enough confidence to classify $x_{lc}$ and it would get most benefits to update.

    \item Margin sampling: Select the samples with smaller margins:
    \begin{equation}
        x_{m} = \arg  \min{P_{C}(\hat{y_1}|x) - P_{C}(\hat{y_2}|x)}
    \end{equation}
    where $\hat{y_1}$ and $\hat{y_2}$ are the first and second most probable class labels under the model, respectively.
    \item Entropy sampling: Select the samples with larger entropy which is an information-theoretic measure that represents the amount of information needed to “encode” a distribution. 
    \begin{equation}
        x_{E} = \arg \max_{x} - \sum{P_{C}(y_i|x)\log{P_C(y_i|x)}}
    \end{equation}

\end{itemize}

\section{System overview}\label{system_overview}

In this section, we provide an overview of our system. 

We proposed a time-series anomaly detection system  RLAD which combines reinforcement learning with active learning.

\tikzstyle{startstop} = [rectangle, rounded corners, minimum width=1.5cm, minimum height=1cm,text centered, text width=1.5cm, draw=black,]
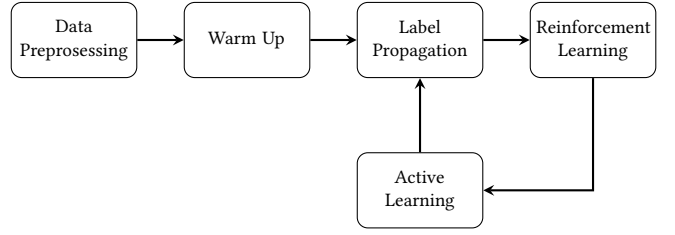
\begin{figure}
\centering
\tikzstyle{arrow} = [thick,->,>=stealth]
{\scalefont{0.8}
\begin{tikzpicture}[node distance=2cm]
\node (start) [startstop] {Data Preprosessing};
\node (pro1) [startstop, right of=start, xshift=0.3cm] {Warm Up};
\node (pro2) [startstop, right of=pro1, xshift=0.3cm] {Label Propagation};
\node (pro3) [startstop, right of=pro2, xshift=0.3cm] { Reinforcement Learning};
\node (pro4) [startstop, below of=pro2] {Active Learning};

\draw [arrow] (start) -- (pro1);
\draw [arrow] (pro1) -- (pro2);
\draw [arrow] (pro2) -- (pro3);
\draw [arrow] (pro3) |- (pro4);
\draw [arrow] (pro4) -- (pro2);

\end{tikzpicture}
}
\caption{Overview of RLAD system}
\label{overview}
\end{figure}

As shown in \ref{overview}, our system contains five modules. Data Pre-processing standardizes values and segments time-series through sliding windows. Also, we initialize label and pseudo label for each timestamp. Then, the preprocessed data will go through Warm Up module. During warming up, an Isolation Forest (iForest)\cite{liu2012isolation} classification model $\Theta_{w}$ is trained. iForest is an unsupervised algorithm which will return anomaly scores for input instances based on $\Theta_{w}$. 

A fraction of instances are used for replay memory initialization. Considering the diversity of replay memory, we select amount of $N_{w}$ instances with highest anomaly scores, $N_{w}$ instances with lowest anomaly scores and $N_{w}$ instances with anomaly scores closed to 0.5. In Label Propagation module,  a label propagation model $\Theta_{lp}$ will be trained by these instances. It propagate labels of these instances and generate pseudo labels for next stage. 

Then, we start training reinforcement learning model $\Theta_{rl}$ based on the true labels and pseudo labels. The Active Learning module will receive the amount of $N_{al}$ selected instances by model $\Theta_{rl}$ based on prediction uncertainty and ask human expert for manually labeling. Again, the new labeled instances are used for label propagation and then threw to training data pool for next iteration training.

\section{Methodology}\label{methodology}
In this section, we describe each component of our approach in detail.  We first introduce the warm up algorithm, followed by the structure of deep reinforcement learning network, label propagation and active learning.

\subsection{Preprossesing and Warm up}
Given a time series sequence $TS = [t_1, t_2, \dots, t_n]$, we separate it into a set of states:
$$S = [(t_1, t_2, \dots, t_{\omega}), (t_2, t_3, \dots, t_{\omega+1}), \dots, (t_{n-\omega}, t_{n-\omega+1}, \dots, t_n)]$$
through a sliding window with size of $\omega$. Each state(sample) has size of $\omega$ and the label of state $S_1$ is determined by the last time point $t_{\omega}$, where 0 represents non-anomaly, 1 represents anomaly and -1 marked as unlabeled. All values are normalized into [0,1] by the \emph{MaxMinScaler}\cite{sklearn}. 

Replay memory is a key component of deep reinforcement learning. It helps the RL agent store some memories and experiences, which speeds up the learning and training. Typically, replay memory stores current transitions(including states, actions and related rewards) with a capacity of $N$. RL agent randomly select a mini batch from replay memory for loss minimization, which will reduce the correlation between samples. However, the normal replay memory has normal distribution and lacks of data diversity. We build a Isolation Forest classification model based on \emph{sklearn}\cite{sklearn} to take a tentative outlier detection. iForest helps us select the most representative anomaly samples and normal samples. We send these samples to RL agent and force it fill the replay memory with transitions of these samples. 

\subsection{Deep reinforcement learning}
For anomaly detection problem, we consider the instance as the state. Action is the prediction of RL agent for whether it is an anomaly or not. $a = 1$ represents anomaly prediction and $a=0$ for non-anomaly prediction. Reward function is set as $(r_1, r_2, -r_1, -r_2)$ for true positives, true negatives, false negatives and false positives.

Typically, policy based reinforcement learning algorithms are used to handle problem with continuous action space. Since our actions are discrete binary values $\in [0,1]$. We select DQN, a value-based algorithm, as our reinforcement learning method.
We adopted a Long Short Term Memory(LSTM) neural network $Eval_{N}$ as the brain of RL agent to generate Q-value when a state $S$ is received, which follows $Q$ function. We build another neural network $Target_{N}$ which has the exactly same architecture with $Eval_{N}$. The later network has lower update rate than the former and we could consider $Target_{N}$ always fixes its parameters. For exploration and exploitation trade-off, an \emph{epsilon decay} strategy is adopted. The greedy factor $\epsilon$ is used to decide whether our RL agent should take actions based on $Q$ function or randomly take actions to explore the entire environment space. The \emph{epsilon decay} policy tries to decrease the percentage dedicated for exploration as time goes by. This can give optimal regret. \cite{exploration}

\begin{algorithm}
  \caption{deep reinforcement learning}
  \KwIn{environment;\\
      set of states $S$;\\
      replay memory $D$ of DQN;\\
      two same structured neural networks $Eval_{N}$ with $Q$ and $Target_{N}$ with $\hat{Q}$;\\
      parameter update ratio $r$;\\
      greedy factor $\epsilon_t$;\\
      discount factor $\gamma$;\\
      learning rate $\alpha$}
  \KwOut{Action set $A$;}
  initial value function $Q$ obtained from Warm Up stage;\\
  
  \For{ $E$ in episodes}
  {
    receive labeled instance set $S$;
    send unlabeled instance set $S_{unlabeled}$ to Active Learning and Label Propagation module;\\
    
    \For{$i$ in epochs}
    { take a state $s$ from $S$;\\
      compute q\_value = $Q(s, a)$;\\
      compute $Prob_{action}$ based on q\_value and $\epsilon$;\\
      action $a$ = random choice with $Prob_{action}$;\\
      observe reward $r$, next state $s'$;\\
      store transition$<s, a, r, s'>$ in $D$;\\
      randomly sample a mini-batch $<s_j, a_j, r_j, s_j'>$from $D$;\\
      target = $r_j +\gamma\max \hat{Q}(s_j, a_j)$;\\
      perform gradient descent: $(q\_value - target)^2$;\\
      \If{$i$ \%  $r == 0$}
      {
        copy parameters to $Target_N$;\
      }
    }
  }
 
\label{algorithm1}
\end{algorithm}

\begin{figure}
    \centering
    \includegraphics[width=\linewidth]{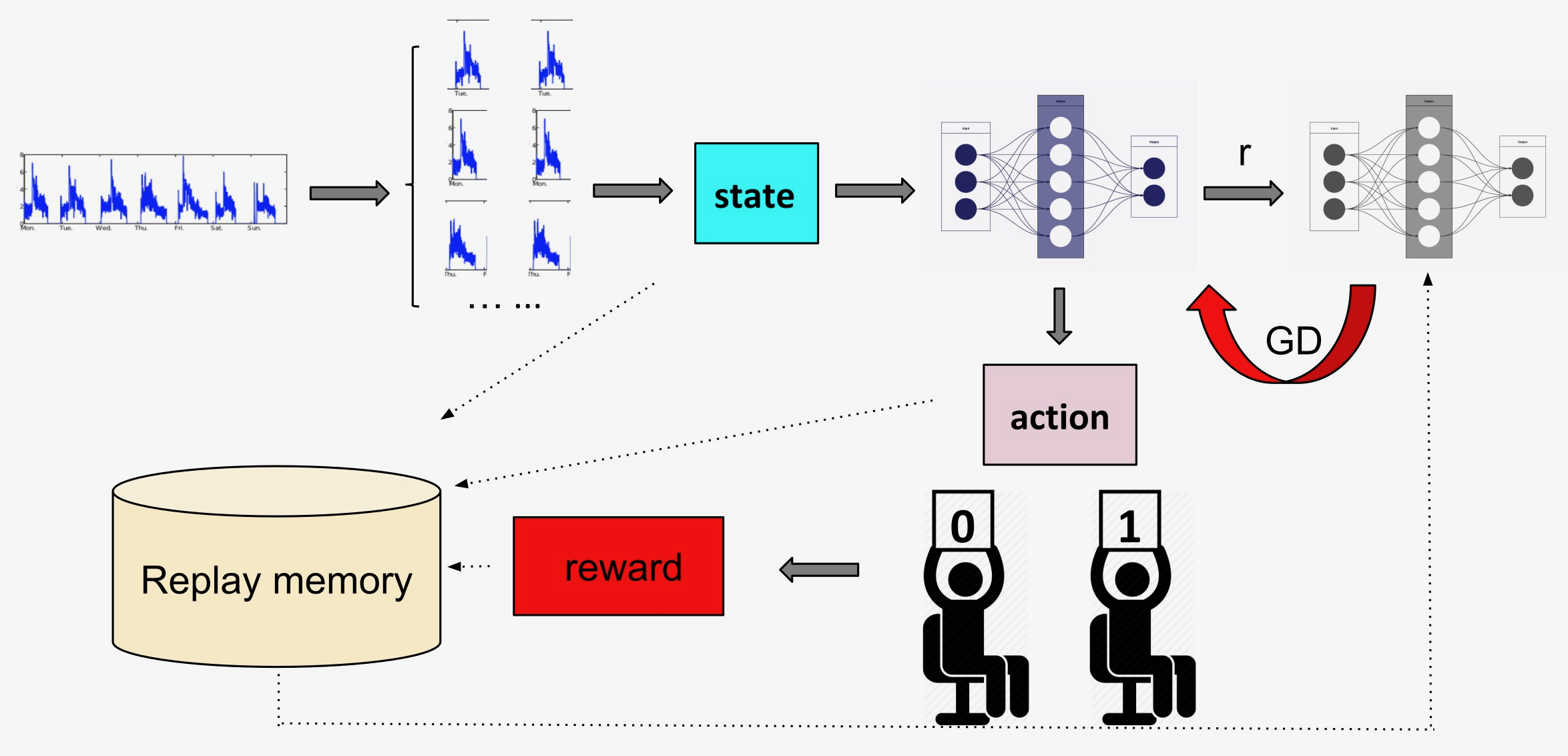}
    \caption{Deep reinforcement learning for anomaly detection}
    \label{fig:rl}
\end{figure}

At the end of Warm up module, the initial reinforcement learning model $\Theta_{rl}$ is trained. As shown in Algorithm 1, for each iteration, DQN receives the set of states $S$ and picks up the labeled ones for training reinforcement learning model $\Theta_{rl}$. The unlabeled instances will be sent to the Active Learning component and Label propagation component which will return a set of new labeled instances for next iteration's training. The transitions will be stored in replay memory. In each epoch, a mini batch of replay memory is sampled and the parameters in $Eval_N$ will be copied to $Target_N$ for every $r$ epochs. A gradient descent is performed for model updating as shown in Figure \ref{fig:rl}.

\subsection{Active learning}
Since fully labeled data is mostly expensive in the real world, we introduced active learning component into our system. It gives our RL agent the capability to not only explore the environment and learn experience, but also ask questions/queries based on its experience during the exploration. 

We choose margin sampling as our active learning strategy. The smaller means our model is more uncertain to identify whether a sample is anomaly or non-anomaly. For each episode, active learning receives the unlabeled instance set $S_{unlabeled}$ from deep reinforcement learning. In each epoch, for a state $s$, the RL agent has two action options: $a_0$ for non-anomaly and $a_1$ for anomaly. Their $q$ values and minimum margin could be formulated as:
\begin{equation}
    (q_0, q_1) = \mathbf{W}* s + \mathbf{b}
\end{equation}
$q$ values reflect the potential reward the RL agent could get when it takes certain action. 
\begin{equation}
    min\_margin = \min |q_0 - q_1| 
    \label{margin}
\end{equation}

We calculate the margin by Equation\ref{margin} and sort them in descending order. Amount of $N_{al}$ with the smallest min\_margin will be inspected by a human expert. We simply assume the expert does not make mistakes and the labels are all correct. Then, the new labeled samples will be sent to propagation and finally thrown into sample pool for next iteration's training. Since this process needs human effort, we count these labels to the total amount labels we used.

\subsection{Label propagation}
We adopted label propagation algorithm to make fully use of labels inspected by human. Label propagation is a semi-supervised algorithm to propagate labels through the dataset along high density areas defined by unlabeled data ~\cite{xiaojin2002learning}. We assume our examples are $X = (x_1, x_2, \dots, x_n)$ with $x_i \in \mbox{\Large$\chi$} $ and the labeled examples are $X_L = (x_1, x_2, \dots, x_l)$ with $x_i$ for $i  \in L=(1,2,\dots,l)$ and the label collection $Y_L = (y_1, y_2, \dots, y_l)$ with $y_i \in C$, where $C$ is the label set for $c$ classes. We define the remaining examples $ X_U = (x_{l+1}, x_{l+2}, \dots, x_{l+u})$ have unobserved label set $Y_U =( y_{l+1}, y_{l+2}, \dots, y_{l+u})$. Thus, the label propagation problem is to estimate $Y_U$ from $X$ and $Y_L$ based on transductive learning.

Label propagation algorithm uses Euclidean distance to measure the similarity between two data points. In a fully connected graph, the weight of edge with parameter $\sigma$ between nodes $i, j$ can be represented by:\\
\begin{equation}
    \omega_{ij} = \exp{\bigg(-\frac{d_{ij}^2}{\sigma^2}\bigg)}
     = \exp{\bigg(-\frac{\sum_{d=1}^D(x_i^d-x_j^d)^2}{\sigma^2}\bigg)}
\end{equation}

The labels could propagate from a node to the neighbor nodes without labels through edges, according to the 
distributions over labels. Edges with larger weights could propagate easier. The probabilistic transition matrix $T$:
\begin{equation}
    T_{ij} = P(j \rightarrow i) = \frac{\omega_{ij}}{ \sum_{k=1}^{l+u} \omega_{kj} }
\end{equation}

And the label matrix $Y$ is defined as its $i$th row representing the probabilistic transition distribution
of node $x_i$. The propagation process begins with $Y \xleftarrow[]{}TY$ and then row-normalize on $Y$ and finally clamp the labeled data and repeat the process until $Y$ converges.

In our system, the label propagator is trained by data with given labels. When a new instance in a time-series is labeled manually, the propagator will generate pseudo-labels to the data instances with similar distributions based on the entropies of transduced label distributions with probability of $pk$:

\begin{equation}
    S = -\sum{(pk * \log{pk})}
\end{equation}

Since these label are generated by label propagation model, we consider them as pseudo-labels and do not count them to the total labels we use.

\section{Experiments} \label{experiment}
In this section, we introduce our experiments in detail. First, we present the data sets we used and then we evaluate our technique and compare with other anomaly detection algorithms. 

\subsection{Dataset}
We used two datasets for our evaluation. Yahoo Benchmark and KPI. They are commonly used for time-series anomaly detection. We provide an overview in Table \ref{dataset} and give a more detailed description then. 

\subsubsection{Yahoo Benchmark}
    Yahoo published its Webscope data set for time-series anomaly detection. It contains real traffic communication of Yahoo services and synthetic curves. In our experiments, we evaluate on both real time series data set \emph{A1Benchmark} which shows the Yahoo membership login data and synthetic one \emph{A2Benchmark} which contains the anomalies of single outliers. \emph{A1Benchmark} includes 67 curves with labels for each timestamp. Each sequence has 1400-1600 data points. \emph{A2Benchmark} has 100 synthetic time series with 1400-1600 points.
    
\begin{table}[]
\centering
\begin{tabular}{|l|l|l|l|}
\hline
             & Yahoo A1 & Yahoo A2 & KPI           \\ \hline
total points & 94866             & 142100            & 3004066       \\ \hline
anomalies    & 1669(1.76\%)      & 400(0.28\%)       & 79554(2.65\%) \\ \hline
\end{tabular}
\caption{Overview of data sets}
\label{dataset}
\end{table}

\subsubsection{KPI data set}
Another one is KPI data set from AIOPs competition. It is collected from several internet companies such as Tencent, ebay and Sogou. There are 3,004,065 data points with timestamps and labels.

\subsection{Metrics}
We evaluate our technique on several aspects:  Accuracy is to judge whether our anomaly detection system could trigger alarms accurately. Also, label demand is also a key factor of time series anomaly detection system since sufficient labels are hard to find in the real world. Also, we examine the time-complexity of our technique.  We select 3 measurements to measure our accuracy: \emph{Precision}, \emph{Recall} and \emph{F1-score}. \emph{Precision} shows how many of alarms triggered by our system are correct anomalies. \emph{Recall} reflects how many anomalies are detected by our system. \emph{F1-score} combines \emph{Precision} and \emph{Recall} together to avoid a trade-off. Their formulas are given as:
\begin{equation}
    Precision = \frac{TP}{TP + FP}
\end{equation}
\begin{equation}
    Recall = \frac{TP}{TP + FN}
\end{equation}
\begin{equation}
    F1-score = 2 * \frac{Precision * Recall}{Precision + Recall}
\end{equation}

\begin{table*}[h!]
\centering
\begin{tabular}{|l|l|l|l|l|l|l|l|}
\hline
\multicolumn{2}{|l|}{\multirow{2}{*}{techniques}}                    & \multicolumn{3}{l|}{Yahoo A1}       & \multicolumn{3}{l|}{Yahoo A2}       \\ \cline{3-8} 
\multicolumn{2}{|l|}{}                                               & F1-score       & precision & recall & F1-score       & precision & recall \\ \hline
\multicolumn{1}{|c|}{\multirow{6}{*}{unsupervised}} & luminol        & 0.177          & 0.261     & 0.258  & 0.277          & 0.314     & 0.266  \\ \cline{2-8} 
\multicolumn{1}{|c|}{}                              & SR-CNN         & 0.264          & 0.174     & 0.540  & 0.057          & 0.048     & 0.069  \\ \cline{2-8} 
\multicolumn{1}{|c|}{}                              & SPOT           & \textbf{0.446} & 0.513     & 0.394  & \textbf{0.691} & 0.53      & 0.991  \\ \cline{2-8} 
\multicolumn{1}{|c|}{}                              & DSPOT          & 0.423          & 0.421     & 0.426  & 0.525          & 0.685     & 0.425  \\ \cline{2-8} 
\multicolumn{1}{|c|}{}                              & DAGMM          & 0.122          & 0.139     & 0.109  & 0.018          & 0.009     & 0.490  \\ \cline{2-8} 
\multicolumn{1}{|c|}{}                              & Autoencoder    & 0.026          & 0.013     & 0.774  & 0.381          & 1.0       & 0.235  \\ \hline
\multirow{6}{*}{semi-supervised}                    & Deep SAD(1\%)  & 0.042          & 0.022     & 0.459  & 0.112          & 0.134     & 0.096  \\
                                                    & Deep SAD(5\%)  & 0.048          & 0.025     & 0.568  & 0.14  & 0.304     & 0.091  \\
                                                    & Deep SAD(10\%) & 0.047          & 0.025     & 0.564  & 0.11           & 0.376     & 0.064  \\ \cline{2-8} 
                                                    & RLAD(1\%)      & 0.708          & 0.652     & 0.781  & 1.0            & 1.0       & 1.0    \\
                                                    & RLAD(5\%)      & 0.752          & 0.71      & 0.8    & 1.0            & 1.0       & 1.0    \\
                                                    & RLAD(10\%)     & \textbf{0.797} & 0.733     & 0.922  & \textbf{1.0}   & 1.0       & 1.0    \\ \hline
\end{tabular}
\caption{Comparison between RLAD with other approaches on Yahoo dataset}
\label{yahoo_table}
\end{table*}

\subsection{Experiment Setup}
We set sliding window size $\omega$ to 25. The LSTM architecture is composed by an input layer, an output layer and a hidden layer. The forget bias is set to 1.0. Size of replay memory $N$ to 1000, the decay ratio of greedy factor $\epsilon$ is $\frac{1}{500000}$, reward constants $r_1 = 5, r_2 = 1$, discount factor $\gamma$ is set to 0.8 for \emph{Yahoo} and 0.98 for KPI. The warm up number $N_w$ is set to 5 and query number for active learning $N_{al}$ is in range of [1,10] for each iteration. We split training and test set as 0.8: 0.2 for our data sets. Since the difference of data balance between these datasets, we examined \emph{Yahoo} dataset with 1, 5 and 10 active queries each episode while 5 and 10 queries for KPI in each episode. During the experiment, RLAD needs 1000 episodes to converge so that we manually labeled 1000(1\%), 5000(5\%) and 10000(10\%) samples during active learning. RLAD converges faster on KPI with 300 episodes so that we used 1500(0.05\%) and 3000(0.1\%) labels.

\subsection{Compare with existing methods}
First, we compare RLAD with state-of-art unsupervised time series anomaly detection techniques. 
\begin{itemize}
    \item Luminol\cite{luminol}: It is a light and configurable Python library which cooperate several statistic anomaly detection algorithms. It does not need any training and could return anomaly scores directly on test data. We implemented Luminol and tested on the portion of 0.2 for each dataset.
    
    \item SPOT and DSPOT \cite{siffer2017anomaly}: They are proposed to detect outliers in streaming univariate time series and able to select threshold automatically. Both of them need a fraction of data for initialization or calibration. In our experiment, we set the portion as same as our split ratio 0.8:0.2. Also, we set risk parameter to $e^{-4}$. 
    
    \item SR-CNN \cite{srcnn}: It generates additional training data with injected fake anomalies to train its neural network and applies to Spectral Residual(SR) for anomaly detection. 
    
    \item DAGMM\cite{zong2018deep}: It utilizes a deep autoencoder to generate a low-dimensional representation and
    reconstruction error for each input data point, which is further fed into a Gaussian Mixture Model (GMM). Then it  jointly optimizes the parameters of the deep autoencoder and the mixture model simultaneously.
    \item Autoencoder\cite{hawkins2002outlier}: It uses replicator neural networks (RNNs) with three hidden layers to provide a measure of
    the outlyingness of data records.
    
\end{itemize}

Then, RLAD is compared with a semi-supervised method Deep-SAD\cite{pang2019deep}. It is able to learn a latent distribution of low entropy for the normal data, with the anomalous distribution having a heavier tailed, higher entropy distribution:
$$
\max _{p(z | x)} \quad \mathcal{I}(X ; Z)+\beta\left(\mathcal{H}\left(Z^{-}\right)-\mathcal{H}\left(Z^{+}\right)\right)
$$
where $Z^+$ is the latent distribution of the normal data and $Z^-$ is the latent distribution of anomalies. $\mathcal{I}(X ; Z)$ represents the mutual information between $X$ and $Z$. We compare Deep-SAD and RLAD on same fraction of labels. Since Deep-SAD throws normal labels and outlier labels separately, we split labels based on the anomaly ratio of dataset. 10 rounds with different seeds are tested and we selected the best result for comparison.

Note that SR-CNN, DAGMM, Autoencoder and Deep-SAD return prediction scores instead of binary prediction. So we applied Bayesian Optimization algorithm to search the threshold which could reach the highest F1-score.

The results on \emph{Yahoo} dataset are shown in Table \ref{yahoo_table}. It is obvious that RLAD outperforms all unsupervised approaches on both A1Benchmark and A2Benchmark. Specifically, with only 1\% labels, our F1-score is 59\% higher than SPOT which performs best over all unsupervised approaches on A1Benchmark and 45\% higher than SPOT on A2Benchmark. With 10\% labels, RLAD is able to achieve nearly 0.8 F1-score on A1Benchmark and 1.0 on A2Benchmark. When compared with semi-supervised approaches, it is evident that RLAD can achieve much better performance than others when given the same number of labels.

As shown in Table \ref{KPI_result}, RLAD is able to achieve 3x  higher f1-score than all of the unsupervised methods on KPI dataset by labeling 0.1\% samples. Moreover,

\begin{table}[]
\centering
\begin{tabular}{|c|l|l|l|l|}
\hline
\multicolumn{2}{|c|}{\multirow{2}{*}{Techniques}}                                              & \multicolumn{3}{c|}{KPI}            \\ \cline{3-5} 
\multicolumn{2}{|c|}{}                                                                         & F1-score       & precision & recall \\ \hline
\multirow{6}{*}{\begin{tabular}[c]{@{}c@{}}un-\\ supervised\end{tabular}}   & Luminol          & 0.032          & 0.111     & 0.063  \\ \cline{2-5} 
                                                                            & SR-CNN           & 0.166          & 0.195     & 0.145  \\ \cline{2-5} 
                                                                            & SPOT             & 0.033          & 0.545     & 0.017  \\ \cline{2-5} 
                                                                            & DSPOT            & 0.049          & 0.025     & 0.911  \\ \cline{2-5} 
                                                                            & DAGMM            & \textbf{0.177} & 0.198     & 0.16   \\ \cline{2-5} 
                                                                            & Autoencoder      & 0.17           & 0.094     & 0.858  \\ \hline
\multirow{4}{*}{\begin{tabular}[c]{@{}c@{}}semi-\\ supervised\end{tabular}} & Deep-SAD(0.05\%) & 0.088          & 0.046     & 0.863  \\
                                                                            & Deep-SAD(0.1\%)  & 0.128          & 0.055     & 0.653  \\ \cline{2-5} 
                                                                            & RLAD(0.05\%)     & 0.709          & 0.681     & 0.897  \\
                                                                            & RLAD(0.1\%)      & \textbf{0.778} & 0.827     & 0.879  \\ \hline
\end{tabular}
\caption{Comparison between RLAD with other approaches on KPI dataset}
\label{KPI_result}
\end{table}

    \label{fig:yahoo_tradeoff}

\section{Conclusion}\label{conclusion}
In this paper, we introduced RLAD for time series anomaly detection. It is the first attempt on anomaly detection area to combine deep reinforcement learning and active learning to reduce reliability of normal pattern assumption and availability of labels. During experiments, RLAD achieved outstanding performance by labeling a tiny fraction of samples on both real and synthetic dataset. Moreover, we outperform the state-of-art both unsupervised and semi-supervised techniques in our experiments. In the future, we plan to explore the possibility for deploying RLAD on real world  and real-time applications.

\bibliographystyle{ACM-Reference-Format}
\bibliography{sample-base}
\end{document}